\documentclass[10pt,twocolumn,letterpaper]{article}

\usepackage{cvpr}              

\usepackage[dvipsnames]{xcolor}

\definecolor{cvprblue}{rgb}{0.21,0.49,0.74}
\usepackage[pagebackref,breaklinks,colorlinks,citecolor=cvprblue]{hyperref}
\usepackage{nicefrac}       
\usepackage{microtype}      
\usepackage{xcolor}         
\usepackage{colortbl}
\usepackage{amsmath}
\usepackage{amssymb}
\usepackage{mathtools}
\usepackage{amsthm}
\usepackage{amsmath}
\usepackage{caption}
\usepackage{subcaption}
\usepackage{enumitem}
\usepackage{makecell}
\usepackage{float}
\usepackage{placeins}
\usepackage{mathtools}
\usepackage{amsthm}
\usepackage{lineno}
\usepackage{epsfig}
\usepackage{comment}
\usepackage{multirow}
\usepackage{adjustbox}
\usepackage{textcomp}
\usepackage{gensymb}
\usepackage{wrapfig}
\usepackage{setspace}
\usepackage{kantlipsum, wrapfig,graphicx}
\usepackage{sidecap, caption}
\usepackage{soul}
\usepackage[ruled,linesnumbered,noend,vlined]{algorithm2e}

\SetCommentSty{mycommfont}

\newcommand{\yun}[1]{{\textcolor{magenta}{\bf [Yun: #1]}}}

\renewcommand{\thefootnote}{*}

\usepackage[capitalize,noabbrev]{cleveref}

\def\paperID{5231} 
\def\confName{CVPR}
\def\confYear{2024}

\title{GDA: Generalized Diffusion for Robust Test-time Adaptation}

\author{Yun-Yun Tsai$^{1*}$,~~Fu-Chen Chen$^{2}$,~~Albert Y. C. Chen$^{2}$,\\
~~Junfeng Yang$^1$,~~Che-Chun Su$^{2}$,~~Min Sun$^{2}$,~~Cheng-Hao Kuo$^{2}$\\
$^1$Columbia University, $^2$Amazon\\
{\tt\small $^1$\{yunyuntsai,junfeng\}@cs.columbia.edu} \\
{\tt\small $^2$\{cfchen,aycchen,ccsu,minnsun,chkuo\}@amazon.com}}

\begin{document}

\maketitle

\def\thefootnote{*}\footnotetext{Work done in Amazon applied scientist internship}
\begin{abstract}






Machine learning models face generalization challenges when exposed to out-of-distribution (OOD) samples with unforeseen distribution shifts.
Recent research reveals that for vision tasks, test-time adaptation employing diffusion models can achieve state-of-the-art accuracy improvements on OOD samples by generating domain-aligned samples without altering the model's weights. Unfortunately, those studies have primarily focused on pixel-level corruptions, thereby lacking the generalization to adapt to a broader range of OOD types.
We introduce Generalized Diffusion Adaptation (GDA), a novel diffusion-based test-time adaptation method robust against diverse OOD types. Specifically, GDA iteratively guides the diffusion by applying a marginal entropy loss derived from the model, in conjunction with style and content preservation losses during the reverse sampling process. 
In other words, GDA considers the model's output behavior and the samples' semantic information as a whole, reducing ambiguity in downstream tasks.
Evaluation across various model architectures and OOD benchmarks indicates that GDA consistently surpasses previous diffusion-based adaptation methods.
Notably, it achieves the highest classification accuracy improvements, ranging from 4.4\% to 5.02\% on ImageNet-C and 2.5\% to 7.4\% on Rendition, Sketch, and Stylized benchmarks. This performance highlights GDA's generalization to a broader range of OOD benchmarks.

\end{abstract}    
\section{Introduction}
\label{sec:intro}












\begin{figure}[t]
    \centering
    \includegraphics[width=0.90\linewidth]{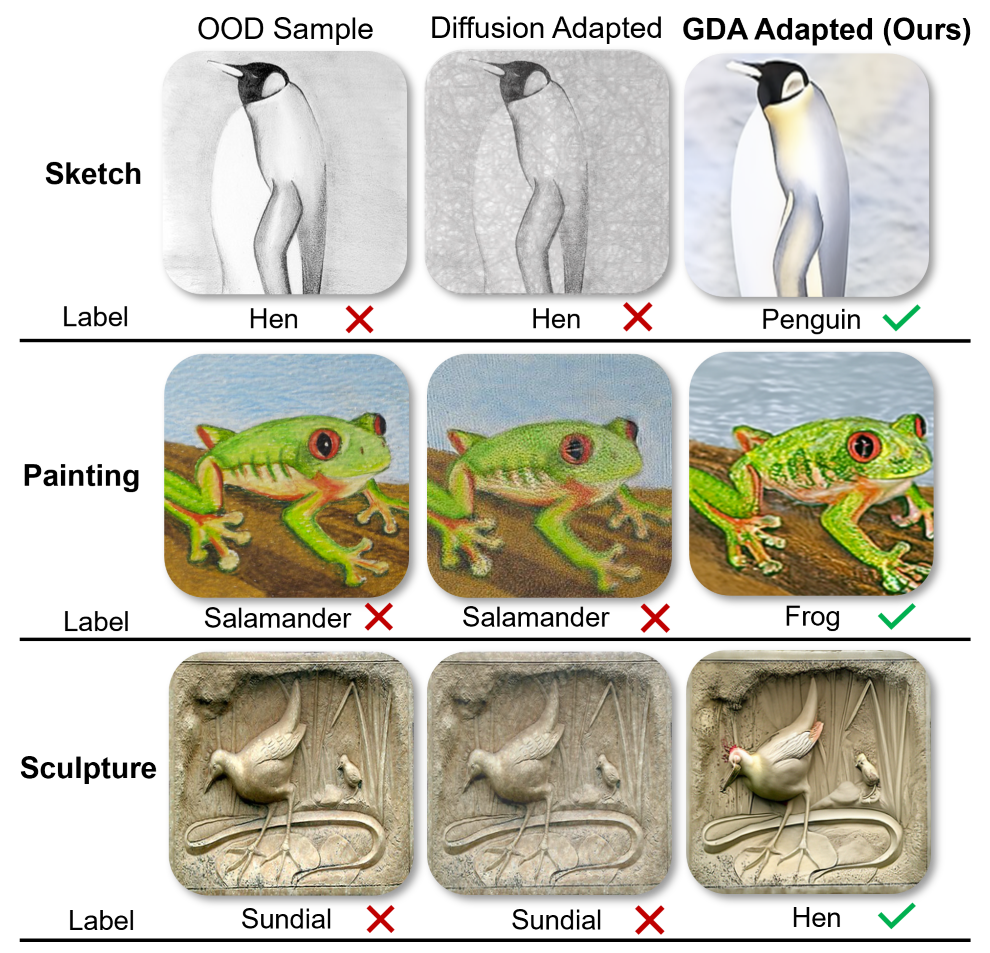}
    \vspace{-1mm}
    \caption{Sample OOD data and adaptations via existing diffusion method and our GDA method.
    The leftmost column shows OOD samples under different style changes, including sketch, painting, and sculpture. The middle column shows samples adapted by traditional diffusion. The rightmost column shows samples adapted with our GDA method. The visualization shows that GDA can generate samples with multiple visual effects, such as re-colorization for the sketch sample, texture enhancement for the painting sample, and object highlighting for the sculpture sample. All three GDA-adapted samples are correctly classified by ResNet50, whereas all others are misclassified.}
    \label{fig:gda_teaser}
    \vspace{-3mm}
\end{figure}

Deep networks have achieved unprecedented performance in many machine learning applications, yet unexpected corruptions and natural shifts at test time~\cite{hendrycks2019benchmarking, hendrycks2019using, hendrycks2019robustness, hendrycks2021many, mao2022causal} still degrade their performance severely.
This vulnerability hinders the deployment of machine learning models in the real world, especially in safety-critical, high-stake applications~\cite{pei2017deepxplore}.

Test-time adaptation (TTA)~\cite{wang2020tent, memo} emerges as a new branch to improve out-of-distribution robustness by adjusting either the model weights or the input data. The former assumes that the weights are not frozen, and can be modified iteratively during test time~\cite{wang2020tent, memo, sun2019testtime}. It thus requires edit access to the model and complicates model maintenance because all adapted model versions need to be tracked. 
The latter modifies the input with random noise vectors or structural visual prompts~\cite{mao2021adversarial, tsai2023self, tsai2020transfer, tsao2024autovp} optimized for pre-defined objectives. The visual prompt design is, however, prone to overfitting due to the high dimensionality of the prompts.

\begin{figure*}[t]
    \centering
    \includegraphics[width=0.96\linewidth]{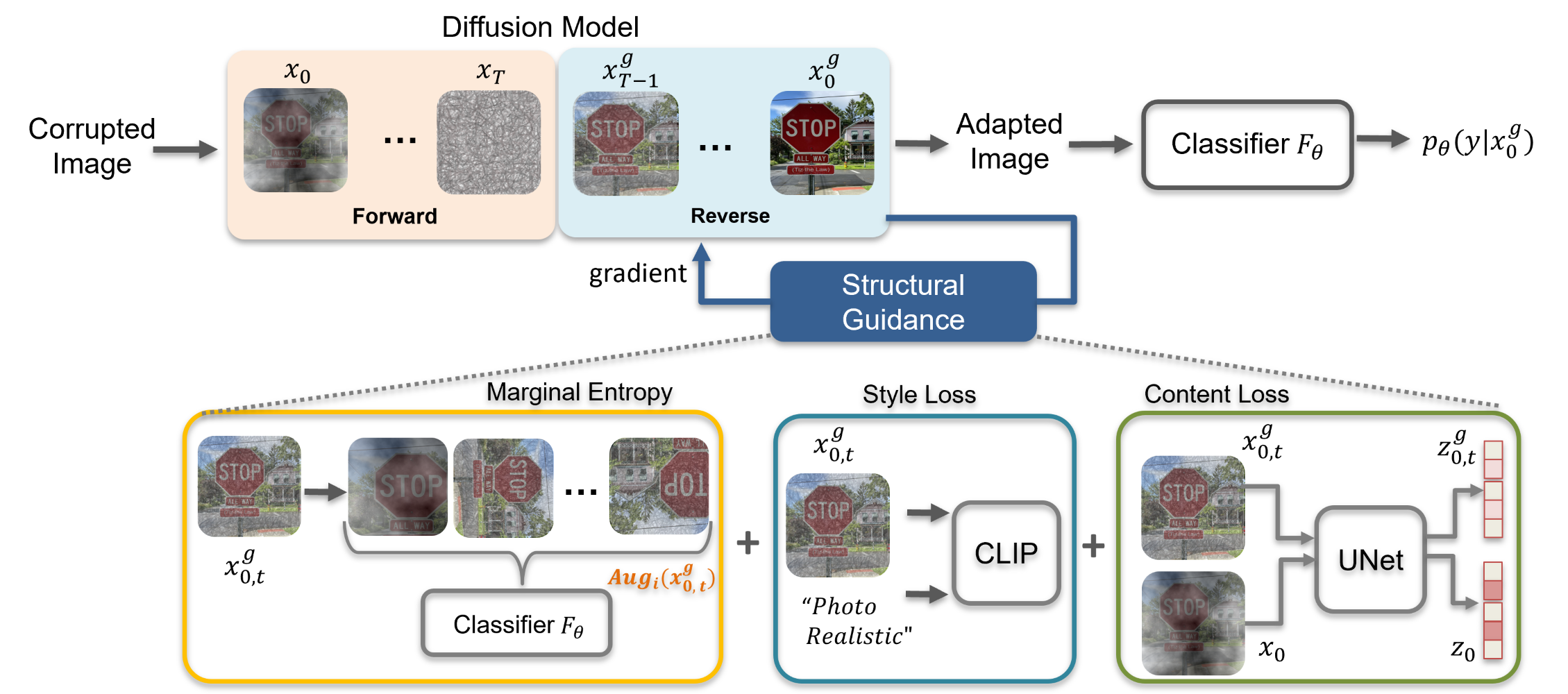}
    \vspace{-1mm}
    \caption{The flow of GDA. We guide the diffusion model with our novel structural guidance that includes marginal entropy, style loss, and content preservation loss. Given the corrupted samples $x_0$, when going through the reverse process at step $t$, our structural guidance will first
    (1) Generate the sample $x^g_{t-1}$ for the next reverse time step $t-1$. (2) Update the $x^g_{t-1}$ with the gradient calculated from the losses. Our loss is computed by the reference image $x_0$ and its corresponding denoised image $\hat{x}^g_{0,t}$ conditioned on $x^g_{t}$ at reverse time step $t$.}
    \label{fig:gda_flow}
    \vspace{-1mm}
\end{figure*}

Therefore, we focus on a new branch of test-time adaptation, diffusion-based adaptation, that does not need to modify model weights and provides more structured guidance.
Prior work~\cite{gao2022back, choi2021ilvr} shows that diffusion is powerful for transferring style and countering natural corruptions by adding simple structural guidance, a latent refinement step conditioned on the input of the reverse process (e.g., a sequence of up-scaling and down-scaling processes). 
However, the key performance gain of prior work~\cite{gao2022back} is shown only in specific corruption types, such as the Gaussian noise or Impulse noise. The results imply two challenges that limit the generalizability of diffusion for adaptation: (1) The structural guidance in prior work can handle only high-frequency corruption and does not generalize well to other types of corruption. (2)
The diffusion model is fully trained on the source domain
data, which potentially causes learning biases and can fail to restore the distribution shift in OOD data.

To address these challenges and improve the generalizability of diffusion models, we propose \emph{Generalized Diffusion Adaptation (GDA)}, an efficient diffusion-based adaptation method robust against diverse OOD shifts at test time, including style changes and multiple corruptions.
Our key idea is a new structural guidance for unconditional diffusion models, consisting of three components: \emph{style transfer}, \emph{content preservation}, and \emph{model output consistency}.
We show sample OOD data adapted by GDA in Fig.~\ref{fig:gda_teaser} and demonstrate the schematic in Fig.~\ref{fig:gda_flow}. To let the corrupted sample shift back to the source domain, GDA incorporates the structural guidance into the reverse process, which has three components: (1) The style loss utilizes CLIP model to transfer the image style; (2) The patch-wise contrastive loss calculated from samples' features aims to preserve the content information; (3) The marginal entropy loss calculated on samples and its augmenting version for ensuring the consistency of output behavior on the downstream task. During the reverse process, GDA iteratively updates the generated samples for every time step by calculating the gradient from three objectives.

The trade-off between style transfer and content preservation in the diffusion model has been studied by~\cite{yang2023zeroshot}. However, the output behavior of the downstream classifier on the generated samples is still unexplored in the diffusion-driven adaptation, which is crucial to the robustness.
Our key insights are: 
(1) Marginal entropy can measure the ambiguity of the unlabeled data with respect to the target classifier~\cite{grandvalet2004semi, memo}.
(2) The marginal entropy calculated from a sample without corruption (clean sample) and its augmented versions is usually lower than a corrupted sample; clean samples are typically less ambiguous to the target classifier.
(3) The diffusion guided with marginal entropy will move the sample away from the decision boundary.




Our main contributions are as follows.
\begin{itemize}
  \item We propose Generalized Diffusion Adaptation (GDA), a new diffusion-based adaptation method that generalizes to multiple local-texture and style-shifting OOD benchmarks, including ImageNet-C, Rendition, Sketch, and Stylized-ImageNet.
  \item 
  Our key innovation is a new structural guidance towards minimizing marginal entropy, style, and content preservation loss. We demonstrate that our guidance is both effective and efficient as GDA reaches higher or on-par accuracy with fewer reverse sampling steps.
\item GDA outperforms state-of-the-art TTA methods, including DDA~\cite{gao2022back} and Diffpure~\cite{nie2022DiffPure} on four datasets with respect to target classifiers of different network backbones (ResNet50~\cite{he2016deep}, ConvNext~\cite{liu2022convnet}, Swin~\cite{liu2021swin}, CLIP~\cite{radford2021learning}).
\item Ablation studies show that GDA indeed minimizes the entropy loss, enhances the corrupted samples, and recovers the correct attention of the target classifier.
\end{itemize}
\vspace{-1mm}


\section{Related Works}
\label{sec:relatedworks}

\subsection{Domain Adaptation}

Various types of out-of-distribution data (OOD) have been widely studied in recent works to show that OOD data can lead to a severe drop in performance for machine learning models~\cite{hendrycks2019benchmarking, hendrycks2021many, recht2019imagenet, mao2022causal, mao2021generative, mao2021discrete}.
To improve the model robustness on OOD data, one can make the training robust by incorporating the potential corruptions or distribution shifts from the target domain into the source domain training data~\cite{hendrycks2021many}. However, anticipating unforeseen corruption at training time is not realistic in practice. 
Domain generalization (DG) aims to adapt the model with OOD samples without knowing the target domain data during training time. 
Existing adaptation methods~\cite{zhou2021domain, dou2019domain, li2018learning, zhou2020deep, memo, sagawa2019distributionally, mao2021generative, mao2021discrete, wang2020tent, sun2019testtime} have shown significant improvement on model robustness for OOD datasets. 

\subsection{Test-time Adaptation}

Test-time adaptation is a new paradigm for robustness to distribution shifting ~\cite{mao2021adversarial, sun2019testtime, memo} by either updating the weights of deep models or updating the input.
BN~\cite{sagawa2019distributionally,li2016revisiting} updates the model using batch normalization statistics. TENT~\cite{sun2019testtime} adapts the model weight by minimizing the conditional entropy on every batch. TTT~\cite{sun2019testtime} attempts to train the model with an auxiliary self-supervision model for rotation prediction and utilize the self-supervised loss to adapt the model. MEMO~\cite{memo} augments a single sample and adapts the model with the marginal entropy of those augmented samples. Test-time transformation ensembling (TTE) ~\cite{perez2021enhancing} augments the image with a fixed set of transformations and aggregates the outputs through averaging. 
Input-based adaptation methods focus on efficient weight tuning~\cite{mao2021adversarial, tsai2023self, jia2022visual, tsai2020transfer, tsao2024autovp} with prompting technique, which modify the pixels of input samples by minimizing the self-supervised loss. Tsai et al.~\cite{tsai2023self} adapt the input by adding a learnable small convolutional kernel and optimizing the parameters during the test time. Mao et al.~\cite{mao2021adversarial} add an additional vector to reverse the adversarial samples by minimizing the contrastive loss. 

\subsection{Diffusion Model for Domain Adaptation}
Recent works have shown diffusion models emerge as a powerful tool to generate synthetic samples~\cite{rombach2022high, song2020improved, pmlr-v139-nichol21a}. A large body of work has studied high-quality image generation by diffusion models. Diffusion models can be widely applied to various computer vision areas, such as super-resolution, segmentation, and video generation~\cite{ho2020denoising, song2020denoising,liu2023accelerating, ho2022video, wang2022semantic}.
In particular, they learn how to reverse the sample from noisy to clean during the training process and the samples are usually drawn from a single source domain. Several works study using diffusion for image purification from out-of-domain data (e.g., corruption or adversarial attack)~\cite{nie2022DiffPure, gao2022back}. Diffpure~\cite{nie2022DiffPure} purifies the adversarial samples by diffusion model by solving the stochastic differential equation (SDE) and calculating the gradient during the reverse process. DDA~\cite{gao2022back} applies diffusion to adapt the OOD samples with multiple corruption types and shows the diffusion-based adaptation is more robust than the model adaptation. However, this approach can only adapt well to noise-type corruption and requires large number of reverse sampling steps (e.g., 50). ILVR~\cite{choi2021ilvr} attempts to
generate diverse samples with image guidance using unconditional diffusion models, but the stochastic nature posed a challenge.
In our work, we investigate how to enlarge the capability of diffusion with a more structured guidance. DSI~\cite{yu2023distribution} improves OOD robustness by linearly transforming the distribution from target to source and filtering samples with the confidence score.
Different from prior works, GDA applies a new structural guidance conditioned on style, content information, and model's output behavior during the sampling process in diffusion models. Our structural guidance is target domain-agnostic, meaning we do not access any ground-truth label or style information of input samples during test time.

\section{Generalized Diffusion Adaptation}
\label{sec:method}

We now introduce our generalized diffusion-based adaptation method (GDA).
Given an unconditional diffusion model pre-trained on the source domain $\mathcal{X}_S$ and an input image $x_0$ sampled from the target domain $\mathcal{X}_T$, the diffusion model should generate samples $\hat{x}_0$ for $x_0$, and the generated samples $\hat{x}_0$ should move closer to the source domain $\mathcal{X}_S$.

We apply the DDPM in our adaptation.
Given an image $x_0$ sampled from the target domain $\mathcal{X}_T$, DDPM first gradually adds Gaussian noise to the data point $x_0$ through a fixed Markov chain during the forward process for $T$ steps. Specifically, we sample data sequence $[x_0, x_1, ... ,x_T]$ by adding Gaussian noise with variance $\beta_t \in (0, 1)$ at timestep $t \in [1, ... , T]$ during the forward process, defined as:

\begin{equation}
   q(x_{t}|x_{0})= \sqrt{\bar{\alpha}_{t}} x_0 +  \sqrt{1-\bar{\alpha}_{t}} \epsilon \ ,
\label{eqn:forward}
\end{equation}
where $\epsilon\sim\mathcal{N}(0, 1)$ is the noise we add, $\alpha_t = 1 - \beta_t$, and $\bar{\alpha}_{t} = \Pi_{s=1}^{t}\alpha_{s}$. 
The reverse process then generates a sequence of denoised image $[x^g_t, x^g_
{t-1} ... ,x^g_0]$ from timestep $t \in [T, ... 1]$. For timestep $t$ in the reverse process, the denoised image can be defined as:
\begin{equation}
    x^g_{t-1} = \frac{1}{\sqrt{\alpha_t}}\left(x^g_t - \frac{1-\alpha_t}{\sqrt{1-\bar{\alpha}_t}}\epsilon_\theta(x^g_t, t)\right)+ \sigma_t \epsilon \ ,
\label{eqn:reverse}
\end{equation}
where $\epsilon_\theta$ is a trainable noise predictor that generates a prediction for the noise at the current timestep and removes the noise. $\sigma_t$ is the variance of noise. Ideally, the generated sample $x^g_0$ should be moved forward to the distribution of the source domain trained for the diffusion model.

\begin{algorithm}[t]
\scriptsize
\DontPrintSemicolon
\SetAlgoLined
\SetNoFillComment
\LinesNotNumbered 
\setstretch{0.99}
	\KwIn{Pretrained classifier $\mathcal{F}(\cdot)$, Augment function set $\mathcal{A}$, OOD images $x_0$, Diffusion time step $T$, Objective function $\ell_{style}(\cdot), \ell_{content}(\cdot),$ and $\ell_{marginal}(\cdot)$, target prompt $r$, Uncertainty score function $H(\cdot)$}
	\KwOut{Class prediction $\hat{y}$ for adapted sample of $x$} 

    \textbf{Inference}\\
    $x_T^g \gets q(x_T|x_0), \quad x_T^g \sim \mathcal{N}(1, 0)$  \tcp*[r]{forward process}
	\For{$t\in\{T,...,1\}$} {
	      $\hat{x}^g_{t-1} = p_\theta(x^g_{t-1}|x^g_t)$ \tcp*[r]{reverse process} 
           $x^g_{0,t} = \sqrt{\frac{1}{\bar{\alpha}_t}}x^g_{t} - \sqrt{\frac{1-\bar{\alpha}_t}{\bar{\alpha}_t}}\epsilon_\theta(x^g_
           {t}, t)$

          $ \ell_{guided} =  \ell_{content}(x^g_{0,t}, x_0)  
           $ \tcp*[r]{structural guidance} 
          $ \quad \quad \quad \quad + \ell_{style}(x^g_{0,t}, r) +  \ell_{marginal}(\mathcal{F}(\mathcal{A}(x^g_{0,t}))) $ 
          
        $ x^g_{t-1} = \hat{x}^g_{t-1} + \bigtriangledown_{x}\ell_{guided}( x)|_{\{x=x^g_{0, t}, x_0\}}$ \\

	}
        \uIf{$H(x^g_{0}) < H(x_0)$}{
          $x^\star \gets x^g_{0}$   \tcp*[r]{confidence filtering}
         }
         \Else{
          $x^\star \gets x_0$ \;
         }
	\KwRet $\hat{y} \gets \mathcal{F}(x^\star)$
\caption{Generalized Diffusion Adaptation}
\label{algo:diffusion_algo}
\end{algorithm}

\paragraph{Structural Guidance in Diffusion Reverse Process}
The trade-off between preserving content while translating domains or style has been studied by DDA~\cite{gao2022back, yang2023zeroshot}. When the noise variance $\sigma$ is more extensive, it is challenging to preserve the content information. Therefore, the structural guidance allows the diffusion model to generate samples conditioned on the predefined objectives. In particular, the structural guidance iteratively refines the latent for the input images during the reverse process so that the content information in the sample can be preserved while translating the style or shifting the domain.





Due to the sampling process of DDPM being a Markov chain, it requires all past denoising steps to obtain the next denoised image. The long stochastic operations can lead to huge distortion of the content information. 
To guide the diffusion more efficiently with structural guidance, we speed up the sampling process with DDIM~\cite{song2020denoising} by skipping several reverse steps.
The reverse process can be redefined as:
\begin{align}
     x^g_{t-1} = &\sqrt{\bar{\alpha}_{t-1}}\left(x^g_t - x^g_{0, t}\right) \\ \nonumber 
     &+ \sqrt{1- \bar{\alpha}_{t-1}-\sigma_t^2}\epsilon_{\theta}(x^g_t, t) + \sigma_t \epsilon \,
\label{eqn:reverse_redefine}
\end{align}
where $x^g_{0,t}$ is the predicted denoised image for $x_0$ conditioned on $x^g_t$ at the time step $t$ and is defined as:
\begin{equation}   
     x^g_{0,t} = \frac{x^g_t - \sqrt{1- \bar{\alpha}_t} \epsilon_{\theta} (x^g_t, t)}{\sqrt{\bar{\alpha}_t}} \ ,
\label{eqn:reverse_denoise}
\end{equation}
Our structural guidance has two steps: (1) At time step $t$, generate the sample $x^g_{t-1}$ for the next step $t-1$. (2) Update $x^g_{t-1}$ with the gradient calculated from our structural guidance.
To avoid the conflicting with the original reverse sampling step at each time step in the diffusion, our structural guidance is computed by the reference image $x_0$ and its corresponding denoised image $x^g_{0,t}$ at reverse time step $t$.
The updated process is defined as:
\vspace{-2mm}
\begin{gather}
    \hat{x}^g_{t-1} \sim p_{\theta}(x^g_{t-1} | x^g_{t})\notag\\ 
    x^g_{t-1} = \hat{x}^g_{t-1} + \bigtriangledown_{x}\ell_{guided}( x)|_{\{x=x^g_{0, t}, x_0\}} \,
\label{eqn:update_sample}
\end{gather}
where $\ell_{guided}$ is our objective function for structural guidance, and the inputs of the objective are $x^g_{0,t}$ and $x_0$. 

\paragraph{Sampling Strategy} 
In Algorithm~\ref{algo:diffusion_algo}, we present GDA. Our proposed structural guidance incorporates the marginal entropy loss into the objective function to ensure the output behavior of the model has consistent predictions on generated samples and their augmented version.
Inspired by~\cite{yang2023zeroshot}, we combine text-driven style transfer using CLIP and content preservation using zero-shot contrastive loss.
Our objective function is:
\begin{equation}
    \ell_{guided}(\cdot) =  \ell_{marginal}(\cdot) + \ell_{style}(\cdot) + \ell_{content}(\cdot) \,
\label{eqn:guided_loss}
\vspace{-1mm}
\end{equation}
where $\ell_{marginal}$ denotes the marginal entropy loss. $\ell_{style}$ and $\ell_{content}$ denote the style and content preservation loss. We further discuss the details for each loss component.


\paragraph{Marginal Entropy Loss}
We notice the stochastic nature of the diffusion model in the reverse process, where the noise $\epsilon$ can lead to the distortion of content information in the input image and cannot correctly generate samples close to the source domain that diffusion has been trained on. 
Given a model $f_{\theta}$ which is trained on the source domain, we add the marginal entropy loss for guiding the diffusion reverse process. In particular, the loss will force the whole diffusion process to generate samples that can decrease the model's uncertainty for $f_{\theta}$. 
At timestep $t$, given a generated sample $x^g_t$ and a set of augmentation functions $\mathcal{A}=\{A_1, A_2, ..., A_n\}$, we augment the sample $x^g_t$ by choosing subset of augmentation functions from $\mathcal{A}$. We denote the image sequence of augmented data as  $A_1(x^g_t), A_2(x^g_t), ..., A_k(x^g_t)$, where $k \leq n$. The marginal output distribution for the given generated sample $x^g_t$ is defined as:
\vspace{-1mm}
\begin{equation}
    \bar{p}_{\theta}(y|x^g_t)  \approx \frac{1}{k} \sum_{i=1}^k p_{\theta}(y|A_i(x^g_t)) \ ,
\label{eqn:update_sample}
\vspace{-1mm}
\end{equation}
where $p_{\theta}$ is the output prediction of each augmented sample and $\bar{p}_{\theta}$ is the average on all augmented samples. Our intuition lies in that $f_{
\theta}$ is trained on the source domain $\mathcal{X}_S=[x_1, x_2, ...,x_N]$ and should learn the invariance between the augmented samples $\hat{x_1}, \hat{x_2}, ..., \hat{x_N}$ and $\mathcal{X}_S$. When generating a sample $x^g_t$ at time step $t \in [T, ..., 1]$ from diffusion, if the sample is close to the source domain, the output prediction of its augmented versions will be consistent, and the marginal entropy loss will become small. Thus, we can utilize this loss to ensure the diffusion process generates samples close to the source domain. Here, the entropy of marginal output distribution is defined as:
\vspace{-2mm}
\begin{equation}
   \ell_{marginal} = - \sum_{y \in \mathcal{Y}} \bar{p}_{\theta}(y|\mathcal{A}(x^g_t)) \log \bar{p}_{\theta}(\mathcal{A}(x^g_t))
\label{eqn:update_sample}
\end{equation}

To better control the sample quality from the diffusion model, the uncertainty estimation on original and adapted samples is then applied to the sampling strategy. The uncertainty score function is $H(x)= - \sum_{y \in \mathcal{Y}} p_{\theta}(y|x)) \log p_{\theta}(x)$, where the input can be the original sample $x$ or adapted samples $x^g_0$.

\paragraph{Style and Content Loss}
To transfer samples from one style to another without content distortion, prior work proposed guided-loss for the diffusion model~\cite{yang2023zeroshot}. Inspired by them,
We use the CLIP model to calculate the style loss. By injecting a text prompt related to the source domain (e.g., \textit{photo-realistic}, \textit{real}), the CLIP model calculates the similarity between the features extracted from the input image and the text prompt. Our style loss is defined as:

\begin{equation}
\ell_{style}=\frac{Enc_{img}(x^g_{0,t}) \cdot Enc_{txt}(t)}{\Vert x^g_{0,t}  \Vert \cdot \Vert t \Vert} \ ,
\label{eqn:style_loss}
\end{equation}
where $Enc_{img}$ and $Enc_{txt}$ are the image and text encoder in the CLIP model.

To avoid content distortion, we use patch-wise contrastive loss to ensure the generated sample's content information is consistent with the original sample. In~\cite{park2020contrastive}, they show contrastive unpaired image-to-image translation loss can preserve the content information by maximizing the mutual information between the input and output patches. To compute the content preservation loss, we extract the spatial features from the UNet component of the diffusion model. The content preservation loss is:
\vspace{-1mm}
\begin{equation}
\ell_{content} = -\log y_{i,j}\frac{\exp(
\hat{z}_i^T z_j)/\tau}{\sum_{k\neq i}\exp(\hat{z}_i^T z_k)/ \tau} \ ,
\label{eqn:content_loss} 
\vspace{-1mm}
\end{equation}
where $\hat{z}$ and $z$ are the corresponding patch-wise features of $x^g_{0,t} 
$ and $x_0$ extracted from UNet $h(\cdot)$. $\tau$ is the temperature scaling value. $y_{i,j}$ is a 0-1 vector for indicating the positive pairs and negative pairs. If $y_{i,j}$ is 1, the $i$-th feature $\hat{z}_i$ and $j$-th feature $z_j$ are at the same location from the $x^g_{0,t}$ and $x_0$ samples. Otherwise, they are from different locations.

\section{Experiment}
\label{sec:experiment}

\vspace{-2mm}

This section presents the details of our experiment settings and evaluates the performance of our method. We comprehensively study multiple types of corruption and style-changed OOD benchmarks. More analyses are shown in Section~\ref{sec:ablation} and Appendix, including sensitivity analysis on different adaptation methods and sample visualization.

\subsection{Experimental Setting}

\paragraph{Dataset.}
We evaluate our method on four kinds of OOD datasets: ImageNet-C~\cite{michaelis2019dragon}, ImageNet-Rendition~\cite{imagenetR}, ImageNet-Sketch~\cite{wang2019learning}, and ImageNet-Stylized~\cite{hendrycks2021nae}. The following describes the details of all datasets.

$\bullet$ \textbf{Natural OOD Data.}  ImageNet-Rendition~\cite{hendrycks2021many} contains 30,000 images collected from Flickr with specific types of ImageNet's 200 object classes. ImageNet-Sketch~\cite{wang2019learning} consists of 50000 sketch images that greatly degrade the performance on large-scale image classifiers.

$\bullet$ \textbf{Synthetic OOD Data.} The corruption data is synthesized with different types of transformations (e.g., snow, brightness, contrast) to simulate real-world corruption. ImageNet-C is the corrupted version of the original ImageNet dataset, including 15 corruption types and five severity levels. 
To evaluate our method, we generate the corruption samples with severity  level 3 based on the official GitHub code~\cite{hendrycks2019robustness}
for each of the 15 corruption types. ImageNet-stylized~\cite{hendrycks2021nae} is another synthetic dataset with huge style change, including eight kinds of styles (e.g., oil painting, sculpture, watercolor, ... etc.). The local textures are heavily distorted, while global object shapes remain (more or less) intact during stylization. We generate the stylized-ImageNet based on the official code~\cite{geirhos2018}


\paragraph{Model.} We use an unconditional 256*256 diffusion model trained with the original ImageNet dataset~\cite{deng2009imagenet}. For the downstream classification models, we test on several architectures, including traditional CNNs, ResNet50~\cite{he2016deep} and ConvNext~\cite{liu2022convnet}; and state-of-the-art transformer Swin~\cite{liu2021swin}. 

\vspace{-2mm}

\paragraph{Baseline Details}
\label{baseline_detail}
We compare our method to several baselines, including standard models without adaption and diffusion-based adaption. 

$\bullet$ \textbf{Standard}: This baseline uses the three pre-trained classification models without adaptation.

$\bullet$ \textbf{DDA~\cite{gao2022back}}: 
This diffusion-based adaptation method provides structural guidance by adding a linear low-pass filter $\mathcal{D}$, a sequence of downsampling and
upsampling operations. We set the reverse step of DDA as 10. The samples will first go through the reverse process and the latent refinement step computes the difference between the output of $\mathcal{D}$ on reference image $x_{0}$ and the generated image.

$\bullet$ \textbf{Diffpure~\cite{nie2022DiffPure}}: 
This baseline uses the diffusion model to purify adversarial samples. It provides an ad-joint method to compute full gradients of the reverse generative process by solving the SDE. Diffpure and DDA
rely on the same unconditional diffusion model but differ in their reverse steps and guidance.

$\bullet$ \textbf{w/o marginal}:
To understand how every objective in our method contributes to the optimization, we remove marginal loss from our method and use only the style and content preservation loss.

\paragraph{Implementation Details}

We adopt the DDPM strategy on the forward and reverse sampling process. 
The total time step $t$ is set as 50. We replace the step size from $T$ to $t$, where $t \in [0, 50]$. Given an input image $x_0$, we obtain the $x_t$ at time step $t$ from the forward diffusion process. 
We combine the three loss terms as a joint optimization, with their Lagrange multipliers as hyperparameters.
The hyperparameter values for each benchmark are shown in Appendix Table~\ref{tab:hyperparameters_weights}.
For the augmentation function $\mathcal{A}$ in marginal entropy loss, we use AugMix~\cite{hendrycks2020augmix}, a data augmentation tool from Pytorch, which randomly select several augmentation functions (e.g., posterize, rotate, equalize) to augment the data.

\begin{table}[t]
\centering
\small
\begin{tabular}{c|ccc}
\hline
                    & \textbf{ResNet50} & \textbf{ConvNext-T} & \textbf{Swin-T} \\ \hline
\textbf{Standard}   & 37.30             & 59.60               & 54.33            \\
\textbf{Diffpure~\cite{nie2022DiffPure}}  & 15.83   & 47.23    & 35.69    \\
\textbf{$DDA_{10}$}~\cite{gao2022back}        & 38.90     & 63.26  & 49.65 \\
\textbf{w/o marg.} & 40.9     & 59.70   & 55.86  \\ 
\textbf{GDA (ours)} & \textbf{41.70}      & \textbf{65.24}       & \textbf{59.35}    \\ \hline
\end{tabular}
\caption{Classification accuracy on the ImageNet-C under severity level 3 for three model architectures. We compare the result between GDA and the four baselines, including Standard, Diffpure~\cite{nie2022DiffPure}, DDA~\cite{gao2022back}, and w/o marginal.  GDA consistently achieves the highest accuracy (numbers in bold) .}
\label{tab:imgnetc_result}
\end{table}

\begin{table}[tb]
\scriptsize
\centering

\begin{tabular}{ccccc}
\hline
\multicolumn{1}{c|}{}           & \textbf{ResNet50} & \textbf{ConvNext-T} & \textbf{Swin-T}  & \textbf{CLIP-B/16} \\ \hline
\multicolumn{5}{c}{\cellcolor[HTML]{C0C0C0}\textbf{Rendition}}                                    \\ \hline
\multicolumn{1}{c|}{\textbf{Standard}}   & 37.0     & 49.8         & 43.6 &  72.7   \\
\multicolumn{1}{c|}{\textbf{Diffpure}}   & 29.8      & 49.4       & 43.5  & 71.4\\
\multicolumn{1}{c|}{\textbf{$DDA_{50}$}}    &   42.0      & 51.8          & 42.1   &  70.6  \\
\multicolumn{1}{c|}{\textbf{w/o marg.}} & 39.4 & 50.5 & 44.2 & 73.4 \\ 
\multicolumn{1}{c|}{\textbf{GDA (ours)}} &    \textbf{44.5}     & \textbf{52.4}          & \textbf{47.6}   & \textbf{76.5}  \\ \hline
\multicolumn{5}{c}{\cellcolor[HTML]{C0C0C0}\textbf{Sketch}}                                       \\ \hline
\multicolumn{1}{c|}{\textbf{Standard}}   & 23.0       & 35.4         & 29.0  &  50.7  \\
\multicolumn{1}{c|}{\textbf{Diffpure}}   &  13.9      & 37.4        & 27.2  & 48.9   \\
\multicolumn{1}{c|}{\textbf{$DDA_{50}$}}    &   23.5     & 34.0          & 27.1  &  44.9  \\
\multicolumn{1}{c|}{\textbf{w/o marg.}}  & 23.9  & 35.7 & 31.1 & 51.2  \\ 
\multicolumn{1}{c|}{\textbf{GDA (ours)}} &    \textbf{25.5}    & \textbf{38.5} & \textbf{35.9}   &  \textbf{55.5} \\ \hline
\multicolumn{5}{c}{\cellcolor[HTML]{C0C0C0}\textbf{Stylized}}                                     \\ \hline
\multicolumn{1}{c|}{\textbf{Standard}}   & 16.5        & 35.3          & 27.3    & 22.4  \\
\multicolumn{1}{c|}{\textbf{Diffpure}}   &  6.1    & 19.8         & 16   & 22.4  \\
\multicolumn{1}{c|}{\textbf{$DDA_{50}$}}    &  19.2       & 27.8          & 18.8   &  21.7 \\
\multicolumn{1}{c|}{\textbf{w/o marg.}} &  20.1       &  36.6        &  30.9 &  22.6\\ 
\multicolumn{1}{c|}{\textbf{GDA (ours)}} &   \textbf{23.0} & \textbf{41.6}          & \textbf{32.3} &  \textbf{25.1}
\\ \hline
\end{tabular}
\caption{The classification accuracy on three OOD benchmarks, including Rendition, Sketch, and Stylized-ImageNet under four model architectures, including ResNet50, ConvNext-T, Swin-T, and CLIP-B/16. We set the timestep for DDA as 50. Numbers in bold show the best accuracy.}

\label{tab:imgnetrss_result}
\vspace{-2mm}

\end{table}

\subsection{Experimental Results}
\begin{figure*}[t]
    \centering
    \begin{subfigure}{.85\textwidth}
      \centering
      \includegraphics[width=.99\linewidth]{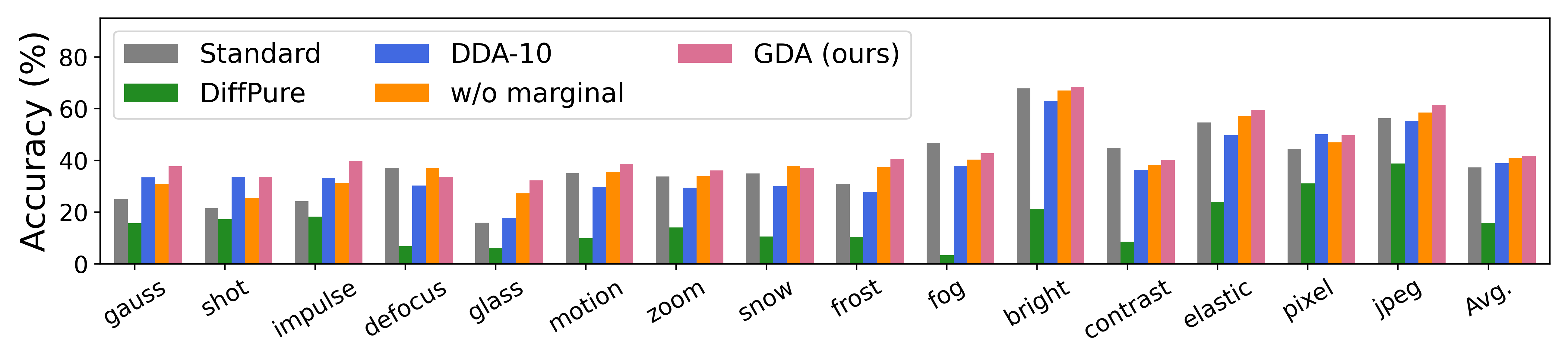}  \vspace{-2mm}
      \caption{ResNet50}
      \label{fig:resnet_corr}
    \end{subfigure}
        \begin{subfigure}{.85\textwidth}
      \centering
      \includegraphics[width=.99\linewidth]{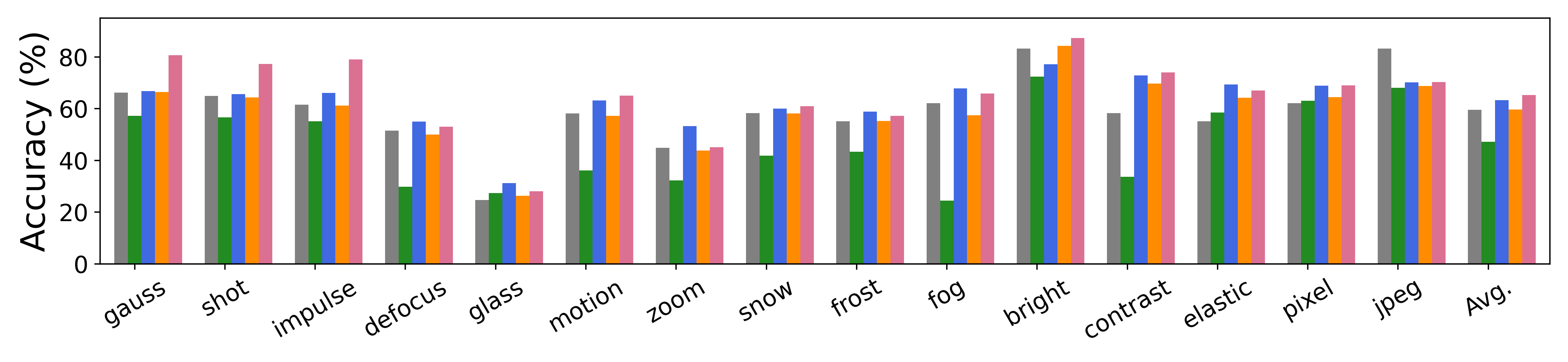}  \vspace{-2mm}
      \caption{ConvNext-T}
      \label{fig:convnext_corr}
    \end{subfigure}
     \begin{subfigure}{.85\textwidth}
      \centering
      \includegraphics[width=.99\linewidth]{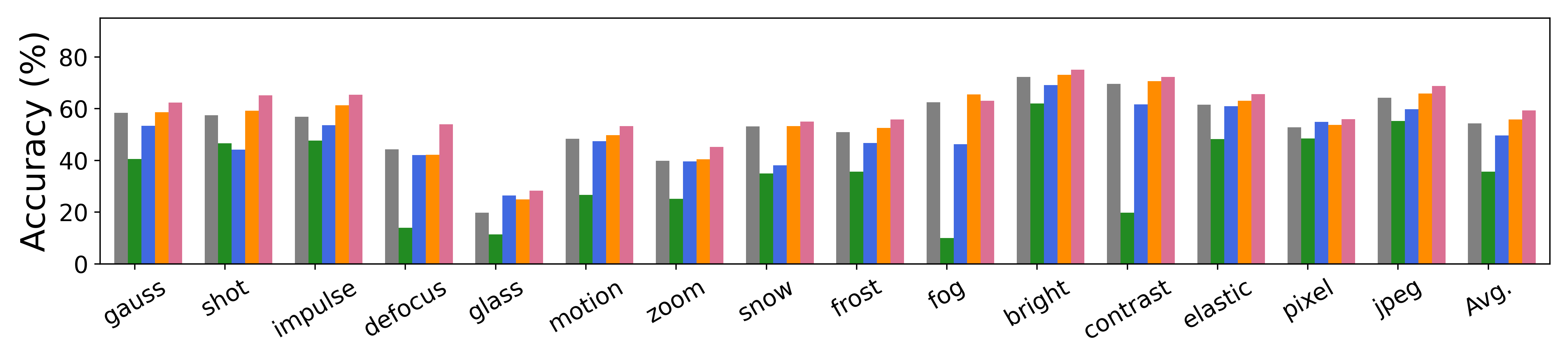} 
      \vspace{-2mm}
      \caption{Swin-T}
      \label{fig:swint_corr}
    \end{subfigure}
    \vspace{-1mm}
    \caption{Comparison of the performance for our method with baselines under 15 types of corruption in ImageNet-C for three model architectures, including ResNet50, ConvNext-T, and Swin-T. GDA shows better improvement on all corruption types for ImageNet-C.}
    \label{fig:corruption_detail_result}

\end{figure*}

Table~\ref{tab:imgnetc_result} shows the results on ImageNet-C. Compared with the three standard models without adaptation, including ResNet50, ConvNext-Tiny, and Swin-Tiny, GDA improves the performance by 4.4\% $\sim$ 5.64\%. Compared with DDA~\cite{gao2022back} and Diffpure~\cite{nie2022DiffPure}, GDA outperforms them by 2 $\sim$ 4\% on average. Besides, to study the effect of marginal entropy, the without marginal
shows the baseline without guiding with the marginal entropy loss. Our results show that the diffusion model can effectively guide the sample back to the source domain with marginal entropy guidance when compared with no marginal guidance and can improve the accuracy by 5.2\%. Fig.~\ref{fig:corruption_detail_result} shows the details of the performance for every 15 corruption types under three model architectures compared with four baselines. In Table~\ref{tab:imgnetrss_result}, we further demonstrate the performance on Rendition, Sketch, and Stylized-ImageNet, which are more challenging datasets with massive style changes. For the Rendition, our method can improve by 2.6$\sim$7.4\% robust accuracy compared with three standard model and outperform state-of-the-art by 0.6\%$\sim$5.5\%. For the Sketch, GDA can improve the accuracy by 2.5\%$\sim$6.9\%. We show the state-of-the-art DDA and Diffpure do not have any improvement on the performance for Sketch dataset. For the Stylized-ImageNet, we improve the accuracy by 6.4\% on average and outperform the state-of-the-art DDA by 2.7$\sim$5\%. In Appendix~\ref{sec:more_exp_results}, we show more experimental results of GDA on ImageNet-C severity 5, and the comparison with other model adaptation baselines.

\begin{table}[t]
\centering
\small
\begin{tabular}{c|cccccc}
\hline
                   &   & \multicolumn{5}{c}{\textbf{\begin{tabular}[c]{@{}c@{}}GDA (ours)\end{tabular}}} \\
\textbf{\# of Aug.}         & 0                                                                       & 2               & 4               & 8               & \textbf{16}              & 32              \\ \hline
\textbf{Rendition} & 39.4                                                                   & 39.7            & 40.5            & 44.2            & \textbf{44.5}            & 44.7            \\
\textbf{Sketch}    & 23.9                                                                    & 22.8            & 23.2            & 24.3            & \textbf{25.5}            & 25.3            \\
\textbf{Stylized}  & 20.1                                                                    & 19.4            & 19.6            & 21.7            & \textbf{23.0}            & 23.5            \\ \hline
\end{tabular}
\caption{The classification accuracy of GDA with different augmentation numbers on Rendition, Sketch, and Stylized-ImageNet OOD benchmarks using ResNet50 model architecture. When number of augmentation is 0, we show the results of GDA w/o marginal guidance. The accuracy values start to saturate when the number of augmentations exceeds 16.
}
\label{tab:num_aug_exp}
\vspace{-4mm}
\end{table}

\paragraph{Number of augmentation in marginal guidance}
In Table~\ref{tab:num_aug_exp}, we show the performance of guiding with marginal entropy loss under different numbers of augmentation on three OOD benchmarks, including Rendition, Sketch, and Stylized-ImageNet. For every step, the marginal entropy loss is computed based on all augmented samples. We set the number of augmentations from 2 to 32. Our result shows that when increasing the number of augmentations to 8 and 16, the performance significantly increases on every benchmark. To be more efficient, in our experiment, we set up the number of augmentation for marginal entropy loss as 16.



\section{Ablation Studies}
\label{sec:ablation}

\begin{figure*}[h]
\centering
    \begin{subfigure}{.33\textwidth}
      \centering
      \includegraphics[width=.99\linewidth]{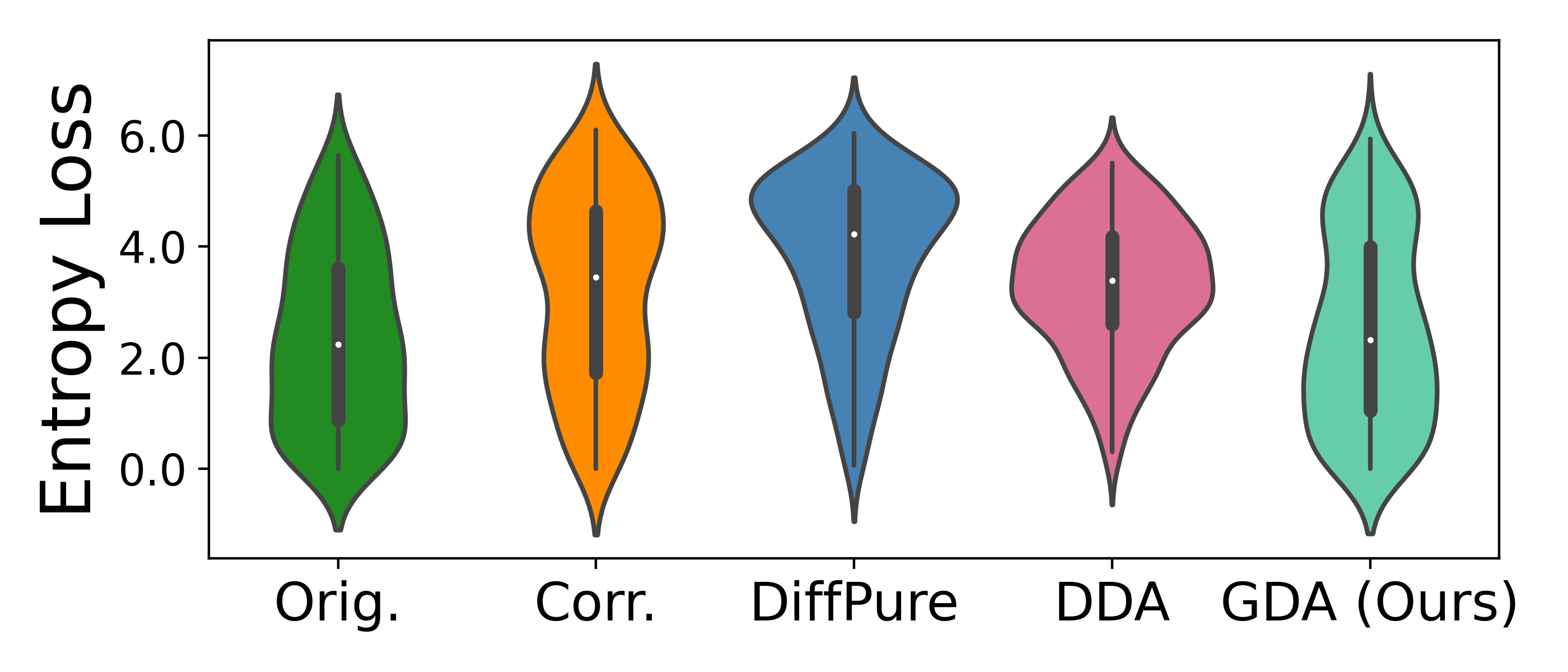}  
      \caption{Frost}
      \label{fig:sub-first}
    \end{subfigure}
    \begin{subfigure}{.33\textwidth}
      \centering
      \includegraphics[width=.99\linewidth]{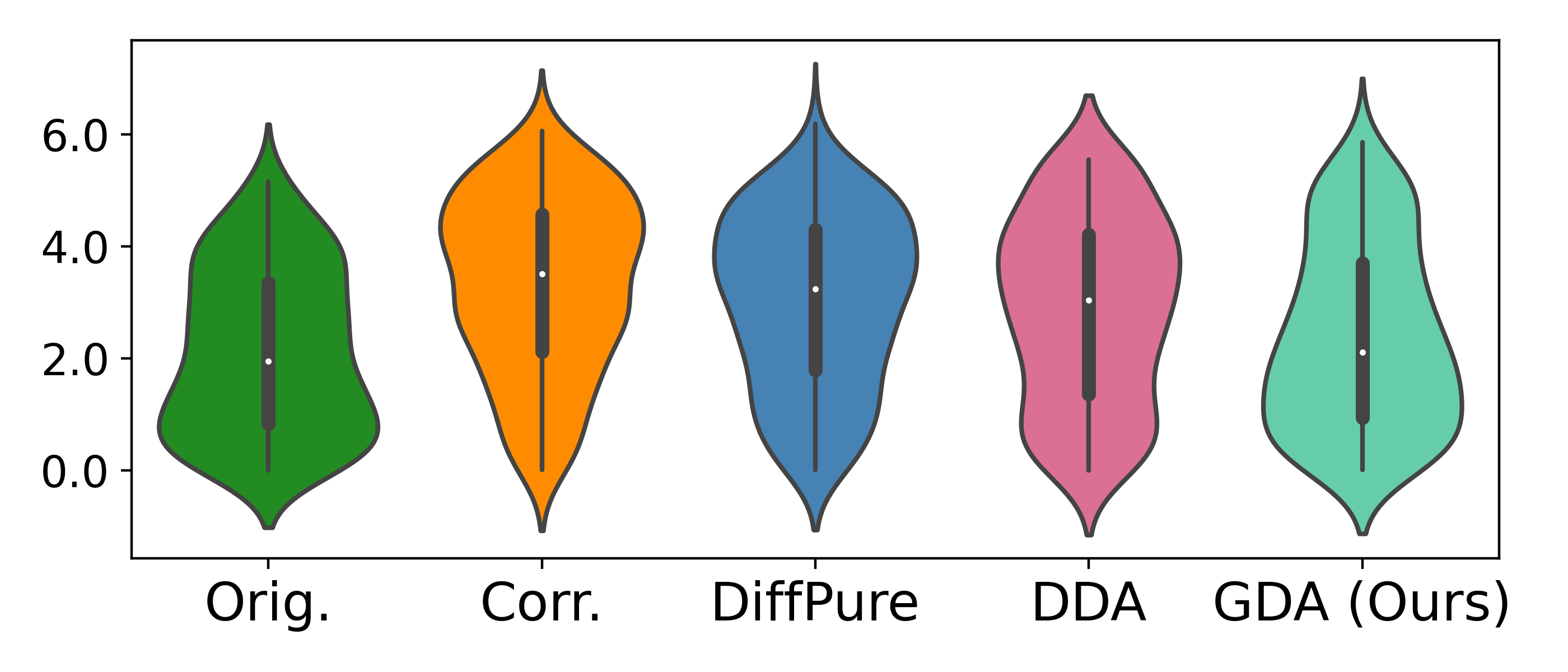}  
      \caption{Gaussian Noise}
      \label{fig:sub-second}
    \end{subfigure}
    \begin{subfigure}{.33\textwidth}
      \centering
      \includegraphics[width=.99\linewidth]{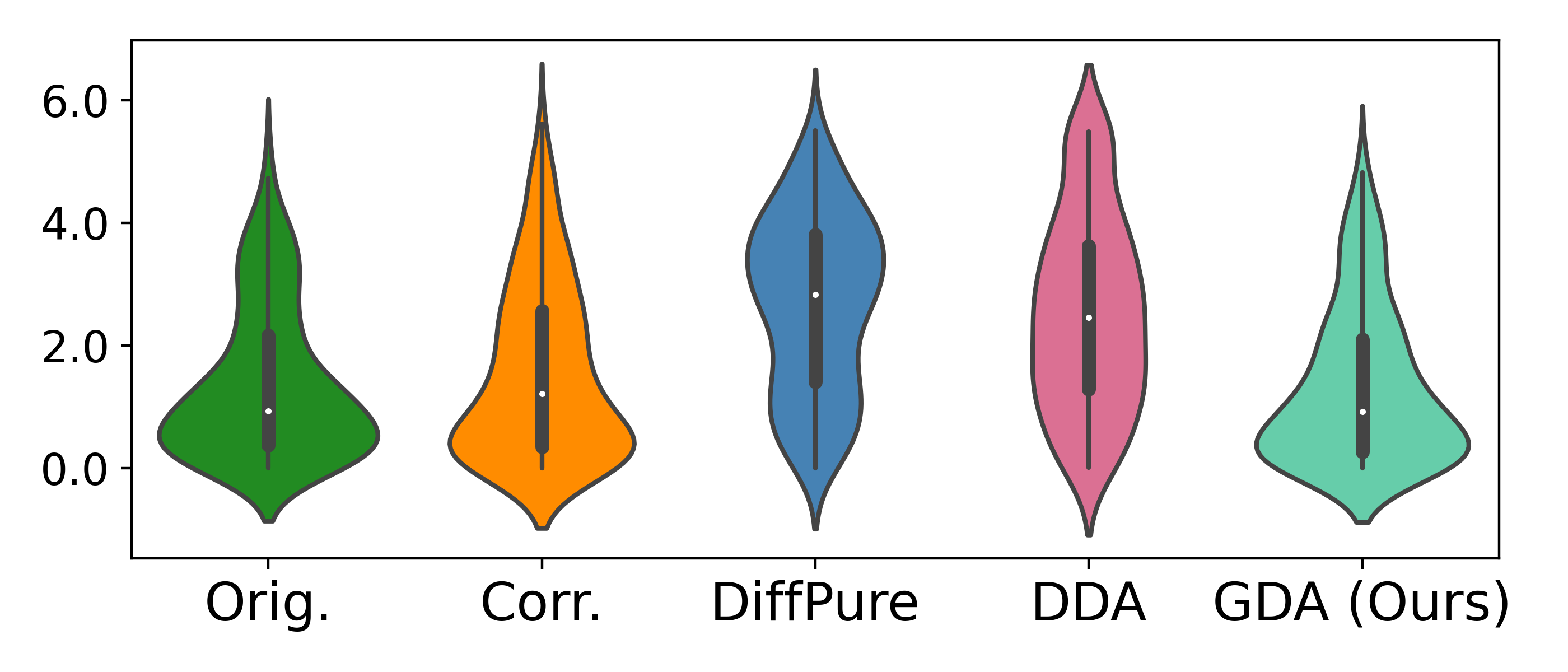}  
      \caption{Pixelate}
      \label{fig:sub-second}
    \end{subfigure}
    \caption{Entropy loss measurement for different corruptions on ImageNet-C. From left to right, the x-axis shows different adaptation methods. The y-axis shows the entropy loss values. The lower value means the model has higher confidence on the sample. In each subfigure, from left to right, we show the loss distribution for original sample (green), corrupted samples (orange), samples adapted by Diffpure~\cite{nie2022DiffPure} (blue), samples adapted by DDA~\cite{gao2022back} (pink), and samples adapted by our method (light green).}
    \label{fig:entropy_violin}
\end{figure*}

 \vspace{-2mm}
\begin{figure*}[t]
\centering
         \includegraphics[width=0.85\textwidth]{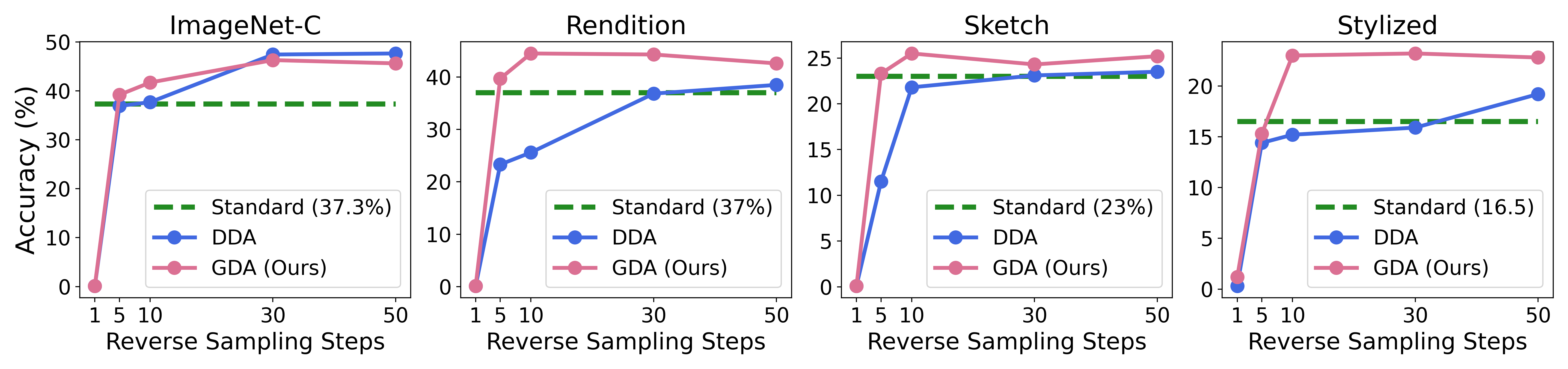}
\caption{Sensitivity analysis on the reverse sampling steps. We compare our method with DDA under different sampling steps from 1 to 50. We evaluate on the ResNet50 model and show the standard accuracy with green color line.}
\label{fig:acc_vs_step}

\end{figure*}

\begin{figure*}[t]
\centering
         \includegraphics[width=0.9\textwidth]{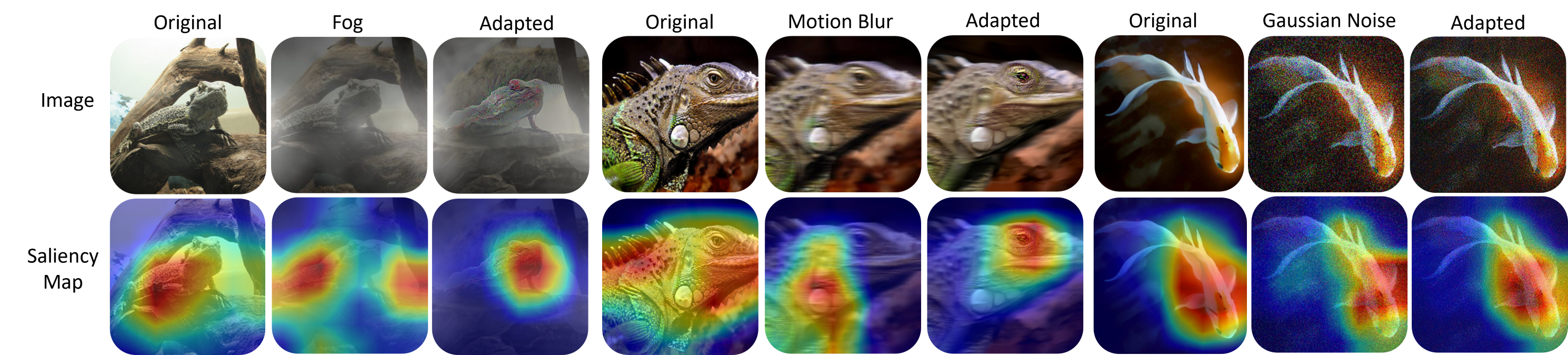}
\caption{GradCam Visualization on ImageNet-Corruption. For every subfigure, from left to right, we show the original, corrupted, and the samples after using GDA to adapt at the first row. The second row shows their corresponding GradCAM. }
\label{fig:gradcam_corr}
\vspace{-2mm}
\end{figure*}


              
      
              

          
       
       


\paragraph{Entropy Loss Measurement}
We do the quantitative measurement of our method by showing the entropy loss distribution for different corruptions. Our entropy is defined as $H(x)= - \sum_{y \in \mathcal{Y}} p_{\theta}(y|x)) \log p_{\theta}(x)$, where it measures the ambiguity of the data with respect to the given target classifier. The lower entropy loss means the model has the higher confidence in the samples. As Fig.~\ref{fig:entropy_violin} shows, the different colors represent different adaptation methods.
The dark green color represents the original sample and the orange color represents the loss distribution of corrupted samples. We show that the entropy loss distribution has a massive shift between corrupted and original samples, which means the model has lower confidence in most of the corrupted samples than the original samples. We then show the entropy loss of samples after adapting with three diffusion-driven adaptation methods, including Diffpure (blue), DDA (red), GDA (light green). As every subfigure in Fig.~\ref{fig:entropy_violin} shows, for every corruption type, the loss distribution of samples generated from GDA moves toward the entropy loss distribution of original samples, which means that our method indeed shifts the OOD samples back to the source domain. However, DDA and Diffpure do not have excessive shifting on the entropy loss distribution.

\vspace{-2mm}

\paragraph{Sensitivity Analysis on Sampling Steps}
In Figure~\ref{fig:acc_vs_step}, we show the effect of different reverse steps on the performance of the diffusion model. In our experimental results in Section~\ref{sec:experiment}, we fix the reverse step number as 10 for every baseline. Here, we compare different reverse sampling steps for DDA and ours from small to large (1 to 50). As Fig.~\ref{fig:acc_vs_step} shows, GDA has a more significant improvement under a small number of reverse steps (e.g., 10) and is more effective compared to the DDA baseline. When increasing the reverse sampling steps to 50, GDA slightly improves but still outperforms the DDA baseline on every OOD benchmark.

\vspace{-2mm}
\paragraph{Analysis on Structural Guidance}

To show how our structural guidance can guide the diffusion model, we visualize the samples generated from GDA and their corresponding gradient classification activation maps (GradCAM). In Fig.~\ref{fig:gradcam_corr}, the corrupted images after adaptation are visually de-corrupted, and the saliency map from GradCAM demonstrates how our objective function can guide the model during the adaptation. In Appendix Fig.~\ref{fig:gradcam_style} and~\ref{fig:gradcam_rendition}, we show the samples from Rendition and Stylized with wrong predictions before adaptation and their corresponding adapted models with correct predictions.

\vspace{-2mm}
\paragraph{Adaptation Cost v.s. Robustness}
In Table~\ref{tab:adaptation_cost}, we show the adaptation cost under different adaptation methods, including DDA, Diffpure, without marginal guidance, and GDA. For GDA, the run time depends on the number of augmented samples. Thus, we select the number with the best accuracy (16) for comparison. Compared to DDA and Diffpure, our method outperforms them by $\sim$7\% on ImageNet-Rendition and reduces 3.85x run time. 


\begin{table}[t]
\centering
\scriptsize

\begin{tabular}{cccccc} \hline
              & Diffpure & $DDA_{10}$ & $DDA_{50}$  & w/o marg. & GDA (Ours) \\ 
Run time       & 31.7 s & 2.1s    & 13.5 s & 2.65 s      & \textbf{3.49 s
} \\
Acc. (\%)    & 29.8  & 24.2   & 42   & 39.4      & \textbf{44.5} \\ \hline
\end{tabular}
\caption{Adaptation run time v.s. Robustness. We show the robust accuracy of Rendition on ResNet50 for every baseline and their corresponding run time for adapting per sample. Compared to DDA and Diffpure, our method outperforms them in smaller run time.}

\label{tab:adaptation_cost}

\vspace{-4mm}
\end{table}



\section{Conclusion}\label{sec:conclusion}
We propose Generalized Diffusion Adaptation (GDA), a novel approach for robust test-time adaptation on OOD samples. 
As opposed to existing methods that require adjusting model weights or inputs with additional vectors, GDA utilizes a diffusion model to shift the OOD samples back to the source domain directly. With our proposed structural guidance based on marginal entropy, style, and content preservation losses, GDA achieves a more generalized adaptation.
Our evaluation results indicate that GDA offers greater robustness across a variety of OOD benchmarks when compared to other diffusion-driven baselines, 
achieving the best accuracy gain on multiple OOD benchmarks.
Our work offers fresh perspectives on OOD robustness by employing the emerging techniques of diffusion models.
For the continued extension of GDA's applications, future research directions include: (1) adapting GDA for tasks such as object detection; (2) investigating a broader range of structural guidance mechanisms, such as incorporating text prompt guidance for the diffusion model; and (3) examining alternative guidance processes to enhance the efficiency of GDA.

\clearpage
{
    \small
    \bibliographystyle{ieeenat_fullname}
    \bibliography{main}
}


\onecolumn
\onehalfspacing

\begin{center}
\Large
\textbf{GDA: Generalized Diffusion for Robust Test-time Adaptation} \\
\vspace{0.5em}
Supplementary Material \\
\vspace{1.0em}
\end{center}

\section{Implementation Details}
\label{sec:implementation_ detail}

\paragraph{Style loss} We apply the CLIP model with model architecture \textit{ViT-Base/16} for calculating the style loss. By leveraging the rich semantic information of CLIP, we are able to shift the OOD sample to the source domain. It has been used in [C2] for style transfer. The input images are presented to the model as a sequence of fixed-size patches, where the patch size is 16*16). We get the corresponding image embedding for all image patches from the output of the visual encoder of CLIP model. We then calculate the similarity between the image embeddings and the text token embedding extracted from language encoder of CLIP model. The text prompts we use for style loss are the words related to \textit{photo-realistic} or \textit{real photo}. 
We assume partially knowing the source domain information is allowable in domain generalization.

\paragraph{Content preservation loss}
We provide a more detailed of the contrastive loss for content preservation. The input of content loss is a batch of features extracted from generated sample itself $x^g_t$ and the corresponding source sample $x_0$. For example, $v$ is the $i^{th}$ patch in sample $x^g_t$, the $i^{th}$ patch $p$ in sample $x_0$ is its positive pair $p+$, and all the other patches except the $i^{th}$ patch in sample $x_0$ will be the negative pair $p-$.
The purpose of the contrastive
loss is to force the feature distance between a patch $p$ and its corresponding positive patch $p+$ to become closer to each other under the latent space. Meanwhile, the loss forces $p$ and $p-$ apart from each other. 

\paragraph{Marginal entropy loss}
We adopt AugMix~\cite{hendrycks2020augmix}, a data augmentation tool from Pytorch, which randomly select several augmentation functions (e.g., posterize, rotate, equalize) to augment the data. The augmention set $\mathcal{A}={A_1, A_2, ..., A_k}$ excludes operations that overlap with corruption types in ImageNet-C. For generating one augmented sample $x_{aug}$, we set the mixing weight $w_1, w_2, ... w_3$ for every augmentation in $\mathcal{A}$. The mixing weight, which is a $k$-dimensional vector of convex coefficients, is randomly sampled from a Dirichlet distribution. The augmented sample $x_{aug}$ equals to  $w_n* A_n(...(W_2* A_2(w1 *A_1(x_{orig}))$.

\paragraph{\textbf{Analysis of Hyperparameters in the Loss Term}} We conduct  the sensitivity analysis on hyperparameters for every loss. We follow the range of hyperparameters used in~\cite{yang2023zeroshot}. In Table~\ref{tab:loss_param_analysis}, we whoe the results of ImageNet-R under different combination of loss terms. 
\begin{table}[h]
\centering
\begin{tabular}{cccccc}
\hline
\multicolumn{6}{c}{\cellcolor[HTML]{C0C0C0}\textbf{Style loss}}    \\
\textbf{Param.} & 1000 & 5000          & 15000 & 20000         & 30000 \\
\textbf{Acc.}   & 38.6 & \textbf{44.5} & 44.0  & 44.2          & 40.1  \\ \hline
\multicolumn{6}{c}{\cellcolor[HTML]{C0C0C0}\textbf{Content loss}} \\
\textbf{Param.} & 100  & 500           & 700   & 1000          & 1500  \\
\textbf{Acc.}   & 38.8 & 39.4          & 42.6  & \textbf{44.5} & 39.8  \\ \hline
\multicolumn{6}{c}{\cellcolor[HTML]{C0C0C0}\textbf{Marginal loss}}  \\   
\textbf{Param.} & 50   & 100           & 150   & 200           & 250   \\
\textbf{Acc.}   & 38.7 & 38.9          & 41.6  & \textbf{44.5} & 42.4  \\ \hline
\end{tabular}
\vspace{-2mm}
\caption{Hyperparameter analysis for ImageNet-R}
\label{tab:loss_param_analysis}
\end{table}

\paragraph{\textbf{The Impact of Different Loss Term}}
We show the impact of different loss term by removing content preservation loss or style loss in Table~\ref{tab:param_analysis2}. The result of using only style loss is better than content loss on ImageNet-Rendition and Sketch.

 \begin{table}[h]
\centering
\begin{tabular}{ccccc}
\hline
                   & \textbf{w/o style} & \textbf{w/o content} & \textbf{w/o marg.} & \textbf{GDA (Ours)} \\
\textbf{Rendition} & 37.7             & 37.9           & 39.4               & 44.5         \\
\textbf{Sketch}    & 23.3             & 23.5           & 23.9               & 25.5         \\ \hline
\end{tabular}
\vspace{-2mm}
\caption{The Impact of Different Loss Term}
\label{tab:param_analysis2}
\end{table}

\paragraph{The Choices of Hyperparameters}
In GDA, the weights for each loss function are hyperparameters that need to be chosen by users. We combine the three loss terms as a joint optimization, with their Lagrange multipliers as hyperparameters.
The hyperparameter values for each benchmark are shown in Table~\ref{tab:hyperparameters_weights}.

\begin{table}[h]
\centering
\begin{tabular}{cccc}
\hline
                    & \textbf{Marg. Entropy} & \textbf{Style} & \textbf{Content} \\
\textbf{ImageNet-C} & 100                    & 5000          & 1500             \\
\textbf{Rendition}  & 200                    & 5000          & 1000             \\
\textbf{Sketch}     & 200                    & 1000          & 700              \\
\textbf{Stylized}   & 200                    & 1000          & 700              \\ \hline
\end{tabular}
\caption{Hyperparmeter setting for marginal entropy loss, style loss, and content preservation loss. The number will be multiplied on every loss function during the optimization.}
\label{tab:hyperparameters_weights}
\end{table}

\clearpage
\section{More Experimental Results}
\label{sec:more_exp_results}
In this section, we show more experimental results on GDA, including the detailed results of ImageNet-C on different severity, comparison with input-based adaptation baselines, and model-based adaptation baselines. 

\subsection{ImageNet-C Detailed Results} 
In main paper Table~\ref{tab:imgnetc_result}, we show the average accuracy on 15 types of corruption for ImageNet-C. Here, in Table~\ref{tab:append_imgnetc},
we show the detailed comparison of GDA with Standard and three diffusion-based baselines. The four main groups of corruption, Noise, Blur, Weather, and Digital, are composed of 15 types of corruptions. We show the detailed corruption types in every group in Table~\ref{tab:corruption_group}. Our GDA improves the robust accuracy by 4.4\%$\sim$5.64\% on three standard models and outperforms every baselines.

\begin{table*}[h]
\centering
\small
\begin{tabular}{clccccc}
\hline
    &      & \textbf{Standard} & \textbf{DiffPure~\cite{nie2022DiffPure}}  & \textbf{DDA-10~\cite{gao2022back}}   & 
          \textbf{w/o marg.} & \textbf{GDA (Ours)} \\ \hline
& Noise     & 23.6   & 17.03     & 33.4   & 29.2       & \textbf{37.0}  \\
& Blur      & 30.5   & 9.28     & 26.8  & 32.4       & \textbf{36.2}  \\
\textbf{ResNet50~\cite{he2016deep}}& Weather   & 45.1  & 11.42       & 39.7 & 46.4       & \textbf{46.5}  \\
& Digital   & 50.1  & 25.62      & 47.9  & 50.9       & \textbf{52.0} \\
& Avg. Acc. & 37.3    & 15.83    & 36.9 & 40.9      & \textbf{41.7} \\ \hline
& Noise     &    64.2  & 56.30   & 66.17  & 63.96 & \textbf{78.99}  \\
& Blur      &    44.83 & 31.4  & \textbf{50.68}   &  44.32    & 47.78 \\
\textbf{ConvNext-T~\cite{liu2022convnet}}& Weather   &   64.67  & 45.46  & 65.92  &   63.75    & \textbf{67.83}  \\
& Digital   &   67.15  & 55.8 & 70.3 &    66.77   & \textbf{70.08} \\ 
& Avg. Acc. &   59.60  & 47.23    & 63.26  &     59.70   & \textbf{65.24}  \\ \hline
& Noise     &  57.56   & 44.93     & 50.4
& 59.7  & \textbf{64.3} \\
& Blur      &   38.05 & 19.27    & 38.85 &  39.3 &  \textbf{45.2} \\
\textbf{Swin-T~\cite{liu2021swin}}& Weather   & 59.68 &  35.63  & 50.05 &  61.1 &  \textbf{62.2} \\
& Digital   &  62.03  & 42.93   & 59.3 &  63.33  & \textbf{65.7}  \\ 
& Avg. Acc. &   54.33  & 35.69   & 49.65  & 55.86 & \textbf{59.35} \\ \hline
\end{tabular}
\caption{Performance on the ImageNet-C for three model architectures under four groups of corruptions. Numbers in bold show the best accuracy.}
\label{tab:append_imgnetc}
\end{table*}

\begin{table}[h]
\centering
\small
\begin{tabular}{c|c}
\hline
                 & \textbf{Corruption Types}                               \\ \hline
\textbf{Noise}   & Gaussian Noise, Impulse noise, Shot noise               \\
\textbf{Blur}    & Motion blur, Zoom blur, Defocus blur, Glass blur        \\
\textbf{Weather} & Snow, Frost, Fog, Brightness                            \\
\textbf{Digital} & Contrast, Jpeg compression, Pixelate, Elastic transform \\ \hline
\end{tabular}
\caption{Detail of four corruption groups with 15 corruption types}
\label{tab:corruption_group}
\end{table}

\paragraph{\textbf{Results of Severity 5}} In Table~\ref{tab:imgnetc_result_s5}, we show more experimental results on ImageNet-C under severity 5.  We compare the results between GDA and the four baselines, including Standard, Diffpure~\cite{nie2022DiffPure}, DDA~\cite{gao2022back}, and w/o marginal. GDA consistently achieves the highest accuracy and surpasses all baselines.

\begin{table}[h]
\centering
\small

\begin{tabular}{c|ccc}
\hline
                    & \textbf{ResNet50} & \textbf{ConvNext-T} & \textbf{Swin-T} \\ \hline
\textbf{Standard}   &  18.7           &  39.3           &   33.1        \\
\textbf{Diffpure~\cite{nie2022DiffPure}}  & 16.8   & 28.8   &   24.8 \\
\textbf{DDA~\cite{gao2022back}}        &   29.7  & 44.2   & 40.0 \\
\textbf{w/o marg.} &  30.2  & 44.4  & 41.6   \\ 
\textbf{GDA (ours)} & \textbf{31.8}      & \textbf{44.8}       & \textbf{42.2}    \\ \hline
\end{tabular}
\caption{The average classification accuracy on the ImageNet-C under severity level 5 for three model architectures.}

\label{tab:imgnetc_result_s5}
\end{table}

\clearpage

\subsection{Compare with Input-based Adaptation}
Similar to our GDA, prior works studied input-based adaptation~\cite{mao2021adversarial, bahng2022visual, tsai2023self}, updating the \emph{input} during the inference time. However, most of them typically focus on adding extra vectors or visual prompts (VP) to the input and optimizing with pre-defined objectives, which is different from our diffusion-based method.
To better understand the efficacy of traditional VP and diffusion-based approaches, we compare the performance of GDA with several input-based adaptation baselines in Table~\ref{tab:vp_baselines}. As Table~\ref{tab:vp_baselines} shows, compared to BN and Memo, GDA outperforms all four input-based adaptation baselines by 2.42\% to 4.46\% in avgerage accuracy, which demonstrates that our proposed diffusion-based method is better than the baselines which add vector directly to the input pixel. 
We explain each input-based adaptation baselines as follows.

\paragraph{Baseline details for input-based adaptation}
\begin{itemize}
    \item \textbf{Self-supervised Visual Prompt (SVP)~\cite{mao2021adversarial}}: The prompting method to reverse the adversarial attacks by modifying adversarial samples with $\ell_p$-norm perturbations, where the perturbations are optimized via the self-supervised contrastive loss. We extend this method with two different prompt settings: \emph{\textbf{patch}} and \emph{\textbf{padding}}. For the patch setup, we directly add a full-size patch of perturbation into the input. For the padding setup, we embed a frame of the perturbation outside the input. 
    \item \textbf{Convolutional Visual Prompt (CVP)~\cite{tsai2023self}}: The prompting method that adapts the input samples by constructing the convolutional kernels. Given a corrupted sample $x$ and a convolutional kernel $k$.
     The convolutional kernels can be initialized with random initialization and optimized with a small kernel size (e.g., 3*3 or 5*5) by projected gradient descent using self-supervised loss.
    We convolve the input $x$ with the convolutional kernel $k$ and update them iteratively by
    $x' = x_0 + \lambda * Conv(x_0, k)$, where the $\lambda$ parameter controls the magnitude of convolved output when combined with the residual input. We set the range to be [0.5, 3] and run test-time optimization to automatically find the optimal solution. We chose the contrastive loss as our self-supervision task.
\end{itemize}

\begin{table}[h]
\centering
\begin{tabular}{ccccccc}
\hline
                 & \textbf{Standard} & \textbf{SVP (patch)} & \textbf{SVP (padding)} & \textbf{CVP (3*3)} & \textbf{CVP (5*5)} & \textbf{GDA}   \\ \hline
\textbf{Noise}   & 28.85             & 29.37               & 29.38                 & 31.59    & 30.53          & \textbf{37.03} \\
\textbf{Blur}    & 30.45             & 29.59               & 29.58                 & 30.80      &31.0        & \textbf{32.4}  \\
\textbf{Weather} & 42.99             & 41.18               & 41.22                 & 42.27         & 42.45     & \textbf{46.5}  \\
\textbf{Digital} & 50.45             & 48.96               & 48.96                 & \textbf{52.58} & 51.45    & 50.98          \\ \hline
\textbf{Avg.}    & 38.19             & 37.27               & 37.28                 & 39.31      & 38.85        & \textbf{41.73} \\ \hline
\end{tabular}
\caption{Compare GDA with input-based adaptation baselines.}
\label{tab:vp_baselines}
\end{table}

\clearpage

\subsection{Compare with Model-based Adaptation}

In Section~\ref{sec:relatedworks}, we introduce prior existing works on \textit{model-based} adaptation, such as TENT~\cite{wang2020tent}, BN~\cite{sagawa2019distributionally}, and MEMO~\cite{memo}. While they all focus on updating the model weights during the inference time, such as changing batch normalization statistics or the scaling parameters in the batch-norm layer, GDA updates the input directly using the diffusion model. We compare our GDA with three model-based adaptation baselines in Table~\ref{tab:tta_baselines}, including TENT, BN, and Memo. For TENT and BN, they adapt the models by input batches, which is different from GDA's setting, as we do the single-sample adaptation. Therefore, we set up the batch size for TENT and BN as 16. For Memo, the same as our single-sample adaptation setting, we set the batch size as 1. We evaluate the accuracy on ResNet50 backbone for every corruption group for GDA and three baselines. As Table~\ref{tab:tta_baselines} shows, compared to BN and Memo, GDA has a 0.3 to 2.7 points gain in robust accuracy. However, GDA is slightly worse than TENT by 2.16 points.

\paragraph{Baseline details for model-based adaptation}
\begin{itemize}
    \item \textbf{BN\cite{sagawa2019distributionally}}: The model adaptation method aims to adjust the BN statistics for every input batch during the test-time. It requires to adapt with single corruption type in every batch.
    \item \textbf{TENT~\cite{wang2021tent}}: The method adapts the model by minimizing the conditional entropy on batches. In our experiment, we evaluate TENT in $episodic$ mode, which means the model parameter is reset to the initial state after every batch adaptation.
     \item \textbf{MEMO~\cite{memo}}: The model adaptation method proposed in ~\cite{memo} alters a single data point with different augmentations (ie., rotation, cropping, and color jitter,...etc), and the model parameters are adapted by minimizing the entropy of the model’s marginal output distribution across those augmented samples.

\end{itemize}

\begin{table}[h]
\centering
\begin{tabular}{cccccc}
\hline
                 & \textbf{Standard} & \textbf{BN~\cite{sagawa2019distributionally}} & \textbf{TENT~\cite{wang2020tent}}  & \textbf{Memo~\cite{memo}}  & \textbf{GDA (Ours)}   \\ \hline
\textbf{Noise}   & 28.85             & 31.14       & 35.75          & 32.61          & \textbf{37.03} \\
\textbf{Blur}    & 30.45             & 28.79       & 33.63          & \textbf{34.31} & 32.4           \\
\textbf{Weather} & 42.99             & 44.81       & \textbf{49.65} & 44.93          & 46.5           \\
\textbf{Digital} & 50.45             & 51.39       & \textbf{56.53} & 53.76          & 50.98          \\ \hline
\textbf{Avg.}    & 38.19             & 39.03       & \textbf{43.89} & 41.40          & 41.73          \\ \hline
\end{tabular}
\caption{Compare GDA with model-based adaptation baselines}
\label{tab:tta_baselines}
\end{table}

\clearpage
\section{Visualization}
\label{sec:visualization}

\begin{figure*}[h]
\centering
         \includegraphics[width=\textwidth]{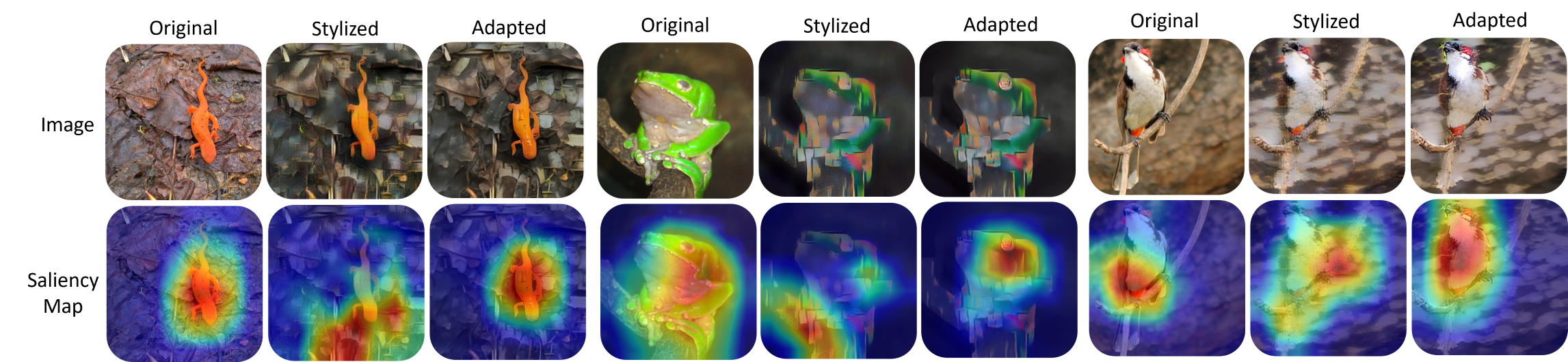}
\caption{GradCam Visualization on ImageNet-Stylized}
\label{fig:gradcam_style}

\end{figure*}

\begin{figure*}[h]
\centering
         \includegraphics[width=\textwidth]{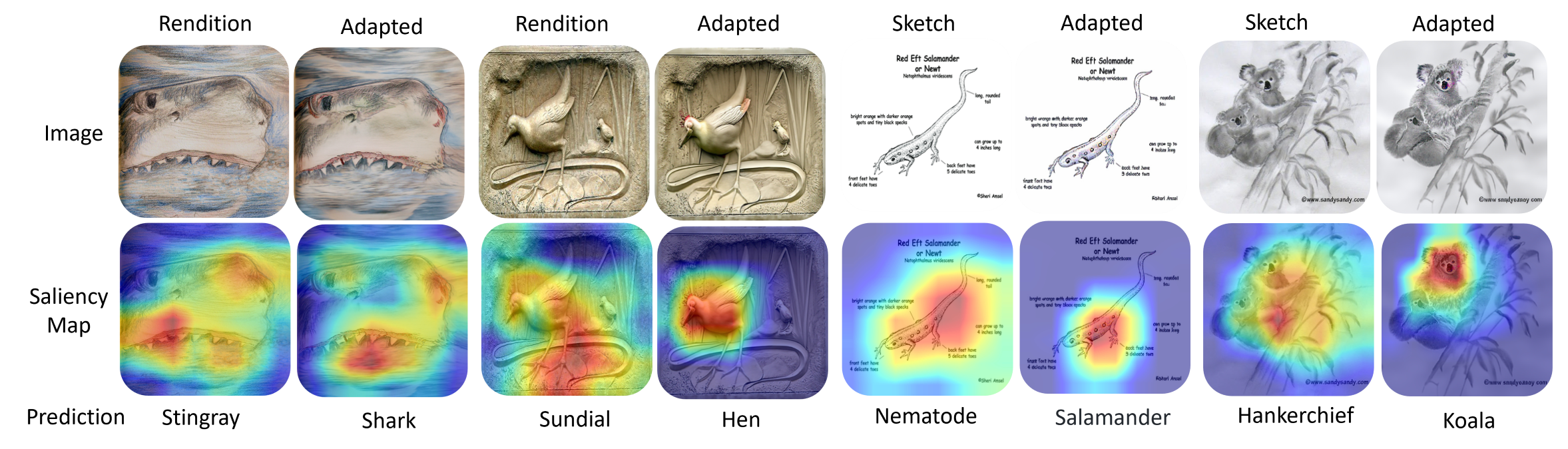}
\caption{GradCam Visualization on ImageNet Rendition and Sketch}
\label{fig:gradcam_rendition}

\end{figure*}

 We visualize more saliency maps on different types of OOD. As Figure~\ref{fig:gradcam_style} and~\ref{fig:gradcam_rendition} shows, from left to right for every subfigure, the first row is the original / corrupted, and adapted samples; the second row shows their corresponding Grad-CAM with respect to the predicted labels. The red region in Grad-CAM shows where the model focuses on for target input. We empirically discover the heap map defocus on the target object for corrupted samples. However, after adapting by GDA, the red region of the adapted sample's heap map is re-target on the similar region as original image, which demonstrates that the diffusion indeed improves the input adaptation and makes the model refocus back on the correct regions.

\begin{figure}[h]
\centering
     \begin{subfigure}[b]{0.65\textwidth}
              
         \includegraphics[width=\textwidth]{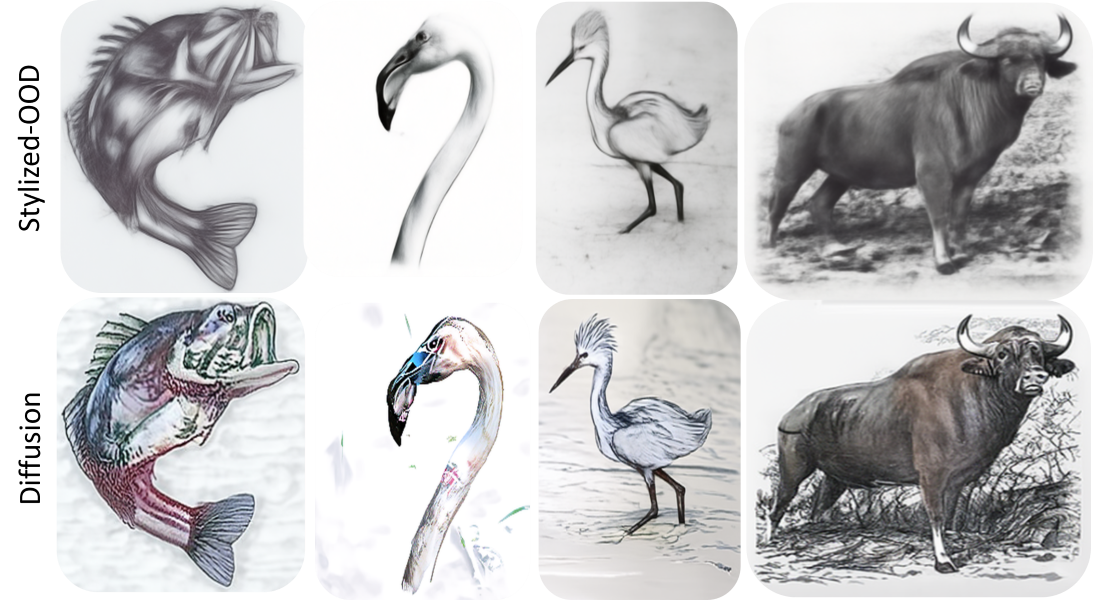}
          \caption{Imagenet-Sketch}
     \end{subfigure}
         \hfill
     \begin{subfigure}[b]{0.65\textwidth}
              
         \includegraphics[width=\textwidth]{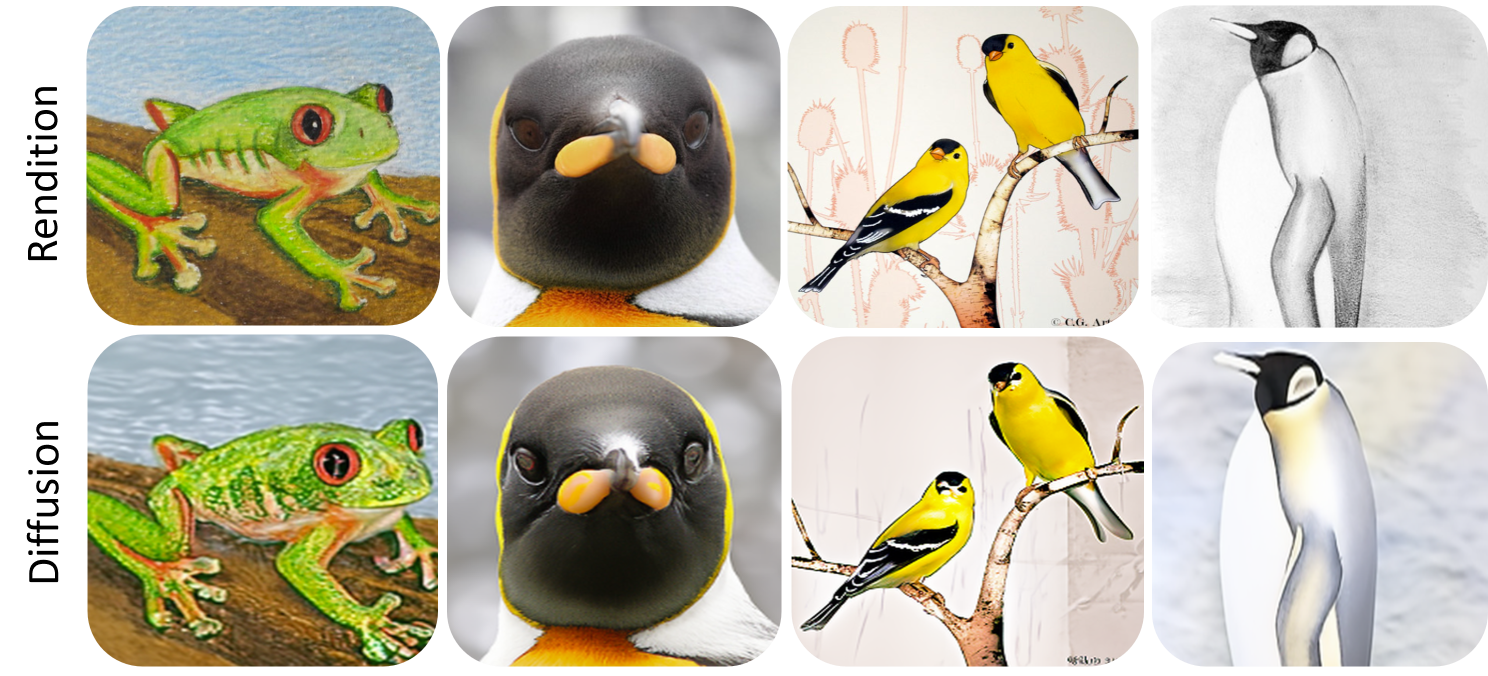}
         \caption{Imagenet-Rendition}
     \end{subfigure}

      \begin{subfigure}[b]{0.65\textwidth}
          
     \includegraphics[width=\textwidth]{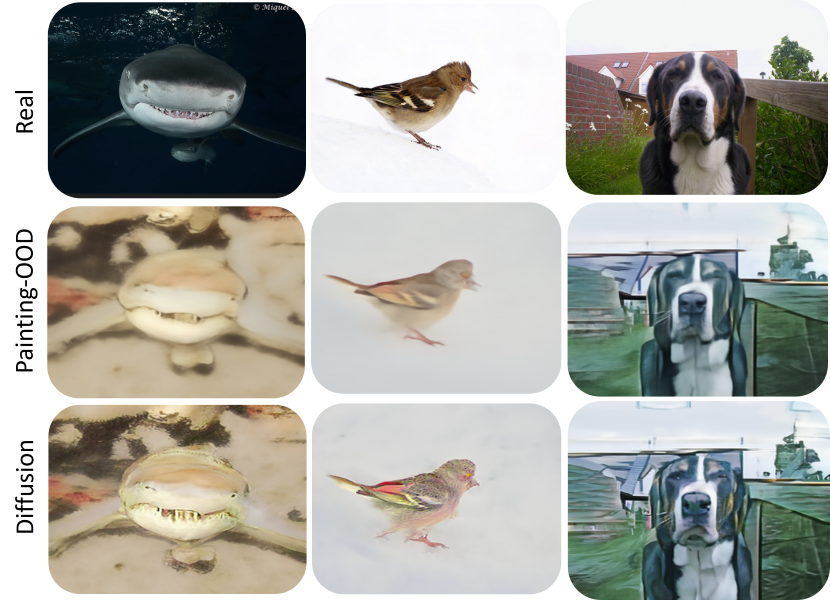}
     \caption{Imagenet-Stylized}
 \end{subfigure}

\caption{More GDA visualization for different OOD benchmarks, including Sketch, Rendition, and Stylized-ImageNet. We show that GDA not only can effectively guide the samples back to the source domain but also can visually change the sample with visual effects, such as colorizing the sketch images, background removing for painting-style samples, and object highlighting for stylized samples.}
\label{fig:sample_vis}

\end{figure}

\end{document}